\documentclass[runningheads]{llncs}
\usepackage{graphicx}

\usepackage{tikz}
\usepackage{comment}
\usepackage{amsmath,amssymb} 
\usepackage{color}
\usepackage{tabularx}
\usepackage{hyperref}

\usepackage[accsupp]{axessibility}  

\def \LTRON{\textbf{LTRON}}
\def \PROBLEM{\textbf{Break and Make}}
\def \PROBLEMLONG{\textbf{Break and Make}}
\def \PHASEA{\textbf{Break}}
\def \PHASEB{\textbf{Make}}

\def \NMODELS{1727}
\def \MINBRICKS{5}
\def \MAXBRICKS{7302}
\def \NBRICKS{1790} 

\begin{document}
\pagestyle{headings}
\mainmatter
\def\ECCVSubNumber{7214}  

\title{Break and Make: Interactive Structural Understanding Using LEGO Bricks} 

\titlerunning{Break and Make: Interactive Structural Understanding Using LEGO Bricks}
%
\author{
Aaron Walsman\and \inst{1}
Muru Zhang\inst{1} \and
Klemen Kotar\inst{2} \and
Karthik Desingh\inst{1} \and \\
Ali Farhadi\inst{1} \and
Dieter Fox\inst{1,3} \\
}
\authorrunning{A. Walsman et al.}
%
\institute{University of Washington \email{awalsman@cs.washington.edu}
\and
Allen Institute for Artificial Intelligence
\and
NVIDIA
}
\maketitle

\begin{abstract}
    Visual understanding of geometric structures with complex spatial relationships is a fundamental component of human intelligence.
    As children, we learn how to reason about structure not only from observation, but also by interacting with the world around us -- by taking things apart and putting them back together again.
    The ability to reason about structure and compositionality allows us to not only build things, but also understand and reverse-engineer complex systems.
    In order to advance research in interactive reasoning for part-based geometric understanding, we propose a challenging new assembly problem using LEGO bricks that we call \PROBLEMLONG{}.
    In this problem an agent is given a LEGO model and attempts to understand its structure by interactively inspecting and disassembling it.
    After this inspection period, the agent must then prove its understanding by rebuilding the model from scratch using low-level action primitives.
    In order to facilitate research on this problem we have built \LTRON, a fully interactive 3D simulator that allows learning agents to assemble, disassemble and manipulate LEGO models.
    We pair this simulator with a new dataset of fan-made LEGO creations that have been uploaded to the internet in order to provide complex scenes containing over a thousand unique brick shapes.
    We take a first step towards solving this problem using sequence-to-sequence models that provide guidance for how to make progress on this challenging problem.
    Our simulator and data are available at \href{https://github.com/aaronwalsman/ltron}{github.com/aaronwalsman/ltron}.
    Additional training code and PyTorch examples are available at \href{https://github.com/aaronwalsman/ltron-torch-eccv22}{github.com/aaronwalsman/ltron-torch-eccv22}.
\end{abstract}

\section{Introduction}

The physical world is made out of objects and parts.
Buildings are made out of roofs, rooms and walls, chairs are made out of seats, backs and legs, and cars have doors, wheels and windshields.
The ability to reason about these parts and the structural relationships between them are a key component of our ability to build tools and shelters, 
solve complex organizational problems and manipulate the world around us.
Building part-based reasoning capability into intelligent agents has been a long-standing goal of the computer vision, robotics and broader AI communities.
\begin{figure*}[t!]
  \centering
  \makebox[\columnwidth][c]{\includegraphics[width=1.0\textwidth]{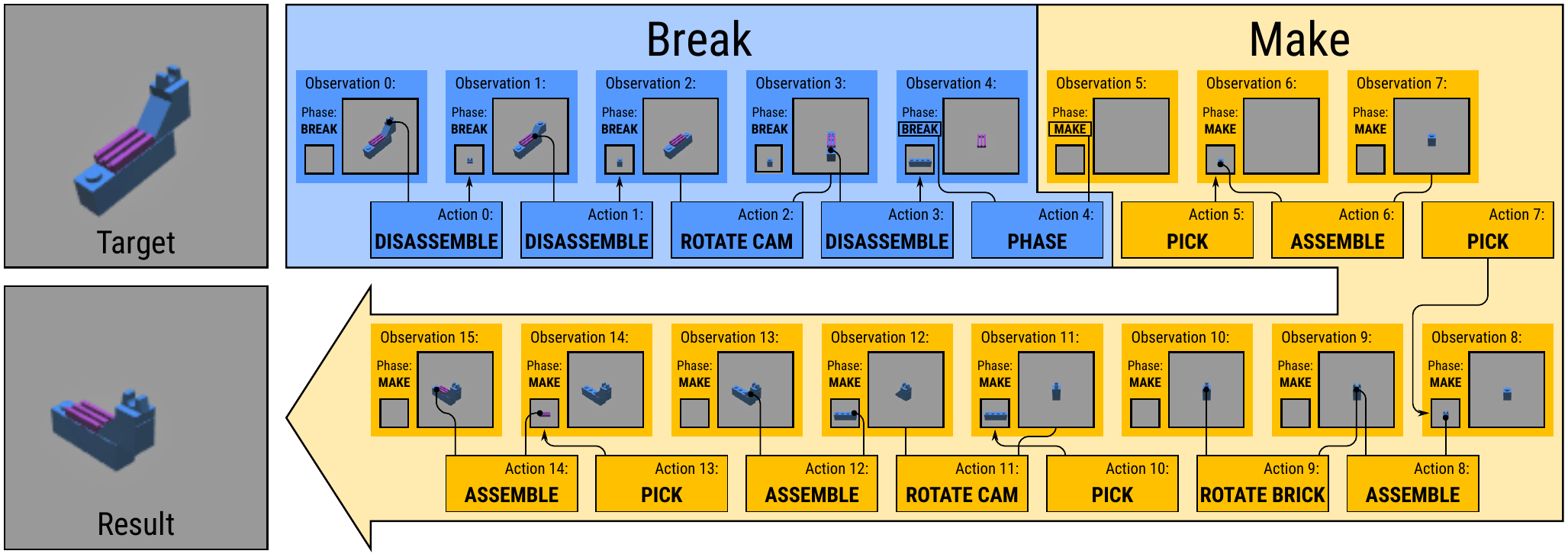}}
  \caption{A training example of the \PROBLEMLONG{} task on a four-brick model in our dataset. During \textbf{Break} phase, the agent must learn to disassemble removable parts based on RGB images to understand the underlying structure. During the second \textbf{Make} phase, the agent must learn to pick bricks and reassemble the scene based on all past observations.
  }
  \label{fig:teaser}
  \vspace{-3mm}
\end{figure*}

In this paper we propose \PROBLEMLONG{}, a challenging new problem designed to investigate interactive structural reasoning using LEGO bricks.  This problem is designed to simulate the process of reverse engineering: taking apart a complex object to learn more about its structure, and then using this newfound knowledge to put it back together again.
This task is naturally divided into two phases.
In the first \PHASEA{} phase, a learning agent is presented with a previously unseen LEGO model and has the opportunity to disassemble and inspect it in order to observe its internal structural and hidden components.  After this, in the second \PHASEB{} phase, the agent is presented with an empty scene and must use the information gathered during the \PHASEA{} phase to rebuild the model from scratch.
Both phases must be completed using visual action primitives designed to simulate the LEGO construction process.
These actions require an agent to reason not only about individual bricks, but also the connection points between them.

In order to facilitate research on this challenging problem, we provide a dataset of \NMODELS{} ethically-sourced fan-made LEGO models with generous public licensing.  These models range in size from \MINBRICKS{} to \MAXBRICKS{} individual bricks and use a library of \NBRICKS{} distinct brick shapes.  We also include a set of augmentations and a random model generator in order to provide more examples for large-scale training.  Finally, we provide a 3D simulator and interactive learning environment with an OpenAI gym interface designed to train agents on this problem.  Our simulator is compatible with a file format commonly used in the LEGO fan community, and is therefore capable of displaying and manipulating a wide range of models found online.

The \PROBLEM{} problem presents a difficult challenge for a number of reasons.
First, the interchangeable nature of LEGO bricks and the large number of distinct brick shapes results in a very large state and action space.
Second, this problem requires precise memory in order to bridge the long-term temporal distance between observations in the \PHASEA{} phase and reconstruction actions that must be taken in the \PHASEB{} phase.  Third, this problem also requires precise spatial reasoning in order to carefully place bricks in the correct location using a visual observation and action space.  Finally, it is difficult to provide direct supervision for this problem, even with accurate information from the simulator.  This stems from the fact that it is not possible to directly compute which observations are necessary to capture the structural details of a model.  We are however able to provide noisy supervision using a custom planner that reasons over visual observations.  This planner generates a series of actions and observations that will feasibly disassemble and reassemble the model, but it comes with no guarantee that an agent equipped with only the visual observations from the sequence would have enough information to make the necessary decisions.

Despite these challenges, we show that progress can be made on LEGO assemblies containing up to 8 bricks using sequence-to-sequence models based on Transformers and LSTMs.  We detail a suite of experiments that show the current capability of these models and demonstrate how performance deteriorates as the problem becomes more complex in terms of both model size and variety of brick shapes.

Our primary contributions are:
\begin{enumerate}
    \item We introduce a challenging new interactive problem \PROBLEM{} requiring complex scene understanding and construction.  Section \ref{sec:data_and_problem} describes this problem in detail.
    
    \item We present \LTRON{}, a new simulator and dataset that allow interactive learning agents to build and manipulate LEGO models.  Sections~\ref{subsec:environment} and \ref{subsec:dataset} provide details.
    
    \item
    We also present a transformer-based network architecture \textbf{StudNet}, designed to make progress on this challenging problem.  We compare this with with other sequence-to-sequence models and demonstrate the difficulty of attacking this problem using current techniques.  Section \ref{sec:method} provides details on these approaches and Section \ref{sec:experiments} discusses their results.
\end{enumerate}

\section{Related Work}
\subsection{Understanding Compositional Structures}
Interactive scene understanding and reasoning about compositional structure has origins in the early days of AI. 
An early example is Winograd's SHRDLU system \cite{shrdlu} that used language instructions to interactively stack virtual blocks and answer questions posed by a human operator.

More recently researchers have introduced a number of interactive environments such as RoboThor\cite{robothor}, iGibson\cite{igibson}, Habitat\cite{szot2021habitat} and MultiON\cite{multion} designed to simulate indoor environments for embodied learning agents.  Many tasks have been proposed for these environments, such as goal-directed navigation~\cite{ZhaoICCV2021},
interactive question answering \cite{iqa,eqa} and instruction following \cite{alfred}.
While many of these tasks and environments offer some degree of object manipulation, most of these interactions involve only a small number of object classes, and do not require the agent to reason about complex compositional structures.  In contrast the \PROBLEM{} task requires an agent to reason in detail about these structures and how to build them from a library of \NBRICKS{} unique parts.

In the non-interactive domain, researchers have released a number of simulated tasks and datasets~\cite{shapenet,mo2019partnet,li2020learning} designed to provide access to a diverse set of objects with increasing detail, part structure and complexity.
Others such as CLEVR~\cite{clevr}, CLEVERER~\cite{clevrer}, and CATER~\cite{cater}
are designed around answering questions about object relationships in images and videos.  In these settings, it is easy to procedurally generate a large dataset using randomization, but it has been difficult to generate datasets with large object and relationship vocabularies. 
Researchers have also taken great effort to annotate natural images and videos with detailed attributes~\cite{farhadi2009describing}, parts~\cite{cub200} and relationships~\cite{visualgenome}.

Scene understanding via active or interactive perception is a classic way for robots and embodied agents to explore and model their environment.
Researchers have investigated varying levels of detail and semantics in this \linebreak space~\cite{vineet2015incremental,mccormac2018fusion++,salas2013slam++,rosinol20203d,sui2020geofusion,nodeslam}.
Previously it has been difficult to explore objects with fine-grained part structure in these settings due to the difficulty in collecting and annotating this data.
\LTRON{} provides complex models in an interactive environment, allowing agents to collect large amounts of data for researching complex cluttered environments with compositional structures.
Another recent line of work explores learning physical properties of the world either from observations of rigid body interactions~\cite{wu2015galileo,wu2016physics,wu2017learning}
or unsupervised physical interaction with a robot~\cite{finn2016unsupervised}.  While we do not provide explicit rigid body dynamics in \LTRON{}, we allow agents to explore extremely detailed physical structures with complex part-interactions 
at a scope that has not been practical in the past.

\subsection{Building 3D Structures}
In robotics, there has been a long-standing interest in enabling robots to build or assemble structured objects. Several authors have explored assembling IKEA furniture~\cite{lim2013parsing,ikea_robot,lee2021ikea}.
Others \cite{inoue2017deep,savarimuthu2017teaching} have used Deep Reinforcement Learning and Learning from Demonstration methods to teach robots high precision assembly tasks using a real robot.  While LTRON{} does not offer the realistic dynamics necessary to support traditional robotic manipulation, it does offer a high degree of scene complexity and compositionality which allows researchers to explore fine-grained spatial reasoning.

Recently construction and object-centric reasoning have become important topics in the reinforcement learning and AI community. Multiple datasets \cite{fusion360,jones2021automate} have been developed to train agents to build and reason about geometric forms using CAD software.  While they support a small number of primitive-based modelling tools, our building environment supports constructing models from over one thousand discrete brick types.   Other recent works \cite{bapst2019structured,ghasemipour2022blocks} have used reinforcement learning for block-stacking problems, and to create structures designed to achieve goals such as connecting or covering other blocks.

Researchers have also investigated the task of generating programs to describe and/or assemble shapes out of low level primitives~\cite{nandi2020synthesizing,jones2020shapeassembly} and reason about the relationships between them~\cite{huang2020generating}.

LEGO bricks are popular construction toys that are often an early entry point for children to learn about building.
They are also an excellent abstraction for real-world construction problems, which has led other researchers to explore using LEGO for various construction problems.
Several approaches have been proposed to automatically construct LEGO assemblies from a reference 3D body \cite{Kim2015SurveyOA}.
For example, multiple authors~\cite{Peysakhov2003UsingAR,lee2015finding} have suggested methods for automated reconstruction based on genetic and evolutionary algorithms.
Duplo bricks have also been used for tracking human demonstrations and assembly~\cite{Gupta:2012:DuploTrack}.

In contrast to these approaches, recent works have suggested data-driven deep learning approaches for LEGO problems based on generative models of graphs \cite{thompson2020building}, and image to voxel reconstruction \cite{lennon2021image2lego}. Similar to \PROBLEM{}, Chung et al. \cite{chung2021brick} propose a method for assembling LEGO structures from a reference image using interactive learning.  Unlike \LTRON{} these approaches use a use only a limited number of bricks, and do not support the large variety of bricks in the LEGO universe.

\section{Task and Data}
\label{sec:data_and_problem}
The \PROBLEM{} task requires an agent to learn how to inspect a LEGO assembly using rendered images, and then use the information gathered in this way to rebuild the assembly from scratch.  Both the inspection phase and the construction phase are inherently interactive problems that require multi-step reasoning due to the ambiguities resulting from occlusions and the iterative nature of the building process.  Many LEGO bricks have groups of similar neighbors which may appear identical under partial occlusion.  Furthermore, complex structures often contain interior bricks that are not visible at all unless outer bricks are removed.
These two factors mean that for many assemblies, there is no single viewpoint that completely captures an entire structure.  Therefore in order to solve this problem an agent must often consider multiple viewpoints and take apart the assembly in order to fully understand it.

\subsection{LEGO Bricks}
A LEGO \textbf{brick} describes the shape and connection-point structure of a single LEGO part.
While most LEGO bricks are a single rigid shape, some such as ropes and connector hoses are flexible.  \LTRON{} currently does not support these flexible components, so they are removed from all models before training.  Some other bricks have moving parts, but in this case we break each of these into a separate brick shape for each moving component.  We use polygon meshes extracted from the LDraw \cite{ldraw} package to represent all bricks.  The \textbf{color} of a brick is represented as a single integer that refers to a specific RGB color value in a lookup table, which is consistent with LDraw conventions.

Each brick also contains a number of \textbf{connection points}.  These describe how bricks may be connected to each other.  The prototypical connection point is the short cylindrical stud that covers the top of many bricks in a rectangular grid, and the corresponding holes that cover the bottom.  However, there are a large number of additional connection point types that exist in the LEGO universe, including technic pins, axles, clips, poles and ball/socket joints.  In developing \LTRON{} we have tried to faithfully represent as many of these as possible in order to provide a rich action space for interactive learning.  Each of these connection points has a number of attributes related to its physical dimensions and compatibility with other bricks.  One important attribute of all connection points is \textbf{polarity}, which describes whether the connection point is an extrusion (positive polarity) or cavity (negative polarity).
We use part metadata from the LDCAD\cite{ldcad} software package in order to detect these connection points on bricks and provide manipulation actions for them.  

We refer to a collection of multiple bricks and their 3D locations as an \textbf{assembly}.  Mathematically, this can be modelled as a set tuples $a = \left\{b_1, b_2, \hdots b_n\right\}$ where each tuple $b_i = (s_i, c_i, R_i, t_i)$ represents an \textbf{instance} of a single brick.  Each of these instances $b_i$ contains a brick shape index $s_i \in N_{shapes}$, a color index $c_i \in N_{colors}$, a 3D rotation $R_i \in SO(3)$ and a 3D translation $t_i \in \mathbb{R}^3$.  The relative placement of the instances, combined with their shapes and the connection points associated with those shapes allow us to construct a set of \textbf{connections} describing a pair of connection points that are in very close proximity to each other and are mutually compatible.

\subsection{Environment}
\label{subsec:environment}
\begin{figure*}[t!]
  \centering
  \makebox[\columnwidth][c]{\includegraphics[width=1.0\textwidth]{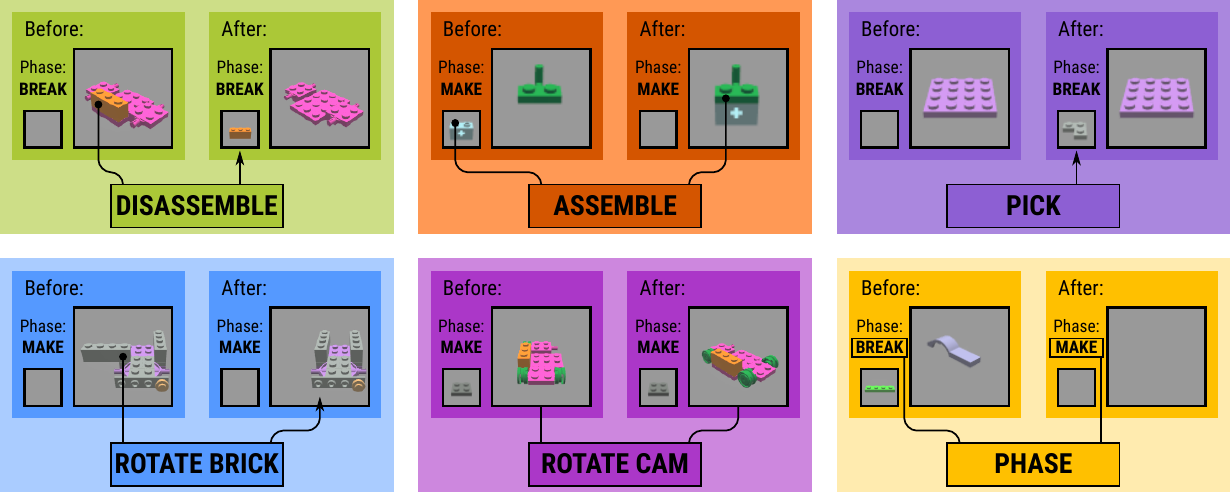}}
  \caption{We define in total six different actions. \textbf{Disassemble} removes a brick from the table workspace by selecting a connection point to detach. \textbf{Assemble} moves a brick from the hand workspace to the scene by attaching a pair of specified connection points. \textbf{Pick} selects a new brick shape and color and adds it to the hand workspace. \textbf{Rotate Brick} rotates the assembled brick at the selected connection point. \textbf{Rotate Camera} rotates the camera to reduce the ambiguity caused by occlusions. \textbf{Switch Phase} switches between break phase and make phase.}
  \label{fig:action_space}
  \vspace{-3mm}
\end{figure*}

In order to manipulate an assembly, our environment provides two virtual work spaces.  The first, which we refer to as the \textbf{table} work space contains the agent's work in progress towards inspecting or assembling a model.  The second work space, which we refer to as the \textbf{hand} contains only a single brick that the agent is about to place, or has just removed from the table workspace.  Each workspace provides a 2D image rendered from a camera viewpoint that can be controlled by the agent.  The table is rendered at $256\times256$ pixels and the hand is rendered at $96\times96$ pixels.
Many of the actions below require the agent to select one or more connection points on bricks in the hand or table workspace.  To do this the agent must specify a 2D location in screen space, and the polarity of the connection point it wishes to select.
This is similar to the Alfred dataset\cite{alfred} and AI2 THOR 2.0\cite{thor2} which allow interaction with objects using pixel-based selection.  To reduce the size of this action space, the resolution of this selection space is downsampled by 4 to $64\times64$ for the table and $24\times24$ for the hand.  \LTRON{} uses these workspaces to provide the following manipulation actions as shown in Fig.\ref{fig:action_space}.

\textbf{Disassemble}: The agent must select a valid connection point in the table workspace.  If the brick can be removed without causing collision, the associated brick instance is removed from the table work space and replaces any brick instance currently in the hand workspace.

\textbf{Assemble}: The agent must specify valid and compatible connection points on one brick in the hand workspace and another in the table workspace.  If the brick may be placed without collision, the brick in the hand is removed and placed into the table workspace attached to the specified connection point.  If the table workspace is empty and there is no destination connection point to select, the agent may select a valid connection point in the hand workspace alone.  This will remove the brick from the hand workspace and add it to the table workspace by placing the specified connection point at the origin.

\textbf{Pick}: The agent specifies a shape id and color id.  A new brick with the specified shape and color replaces any brick instance currently in the hand work space.

\textbf{Rotate 90/180/270}:  The agent must select a valid connection point on a brick in the table workspace.  If rotating the brick will not cause collision, the brick is rotated by the specified angle about the primary axis of the connection point.

\textbf{Rotate Camera Left/Right/Up/Down/Frame}:
In some cases it may be necessary to view an assembly from different viewpoints in order to effectively manipulate it, so we provide five actions for each workspace that the agent can use to manipulate the camera.  The first four rotate the camera up, down, left or right about a fixed center point.  Rotating left and right rotates by 45 degrees about the scene's up-axis, while rotating up and down alternate between a downward viewing angle 30 degrees above the center point and an upward viewing angle 30 degrees below the center point.  The fifth \textbf{Frame} camera action moves the camera's fixed center point to the centroid of the current brick assembly.

\textbf{Switch Phase}:
Finally there are two additional actions that switch from the Break phase to the Make phase, and that end the episode when the agent is finished building.  Switching from the Break phase to the Make phase clears both workspaces.

\subsection{Evaluation}
\label{subsec:evaluation}
The \PROBLEM{} task requires a learning agent to visually inspect a LEGO assembly in order to gather enough information to then build it again from scratch.
In order to assess the capability of a learned model, it is necessary to compare the generated assembly that it builds with the target assembly it is trying to copy.  We provide four different metrics that attempt to estimate various aspects of the agent's success.

\textbf{$\mathbf{F1_b}$ score: }The first metric is an F1 score over bricks in the two assemblies which we refer to as $F1_b$.  This metric ignores pose and simply measures whether the agent was able to add the correct bricks to its estimated assembly regardless of how they are connected together.  For this metric, we first remove pose information from the generated assembly $\hat{a}$ and the target assembly $a^*$ to produce a multi-set of brick shape and colors $m^* = \left\{(s^*_0, c^*_0) \hdots (s^*_n, c^*_n)  \right\}$ for the target assembly and another $\hat{m} = \left\{(\hat{s}_0, \hat{c}_0) \hdots (\hat{s}_n, \hat{c}_n)  \right\}$ for the assembly the agent generated.  We can the compute true positives, false positives and false negatives as:
\[
TP_b = m^* \cap \hat{m},\ \ \ \ FP_b = \hat{m} - m^*,\ \ \ \ FN_b = m^* - \hat{m}
\]
We then use these three quantities to compute an F1 score.
Getting a score of 1.0 on this problem is necessary to rebuilding the assembly correctly, but it is not sufficient.  This metric is still useful though because it allows us to categorize errors.  If the agent was not able to rebuild the structure, but was able to identify the necessary bricks for that structure, then it may give us guidance for which aspect of the system needs the most improvement.

\textbf{$\mathbf{F1_a}$ score: }
Unlike $F1_b$,
$F1_a$ includes pose and is designed to measure the accuracy if the entire assembly.  In this metric, we first define a rotation threshold $\theta_\epsilon$ and a distance threshold $d_\epsilon$ and say that two bricks $i$ and $j$ are \textit{aligned} iff they have the same shape $s_i = s_j$ and color $c_i = c_j$ and their centers are close $||t_i - t_j|| < d_\epsilon$ and the geodesic distance between their orientations is close $G(R_i, R_j) < \theta_\epsilon$.

Given that we care more about the \textit{relative} position of bricks to each other, than their \textit{absolute} position in the scene, we first compute a single rotation $R_0$ and translation $t_0$ that bring as many bricks in $a^*$ into alignment with $\hat{a}$ as possible.  We then consider each brick in $\hat{a}$ to be a true positive if it is aligned with another brick in $a^*$ and consider it to be a false negative otherwise.  Any brick in $a^*$ that is not aligned to a brick in $\hat{a}$ is a false negative.  We then use these quantities to compute $F1_a$.

\textbf{Assembly Edit Distance (AED): }While this $F1_a$ metric gives us a useful measure of similarity between two assemblies, it is possible that it may over-penalize some small mistakes.  Consider the case where a long chain of bricks has been reconstructed correctly except for a single mistake in the middle.  Because of the single rigid transform $R_0$ and $t_0$, we can only align either the top half or the bottom half of the reconstruction $\hat{a}$ with the target assembly $a^*$, and will incur a massive penalty for this single mistake. 
To mitigate this, we introduce Assembly Edit Distance (AED): we compute $R_0$ and $t_0$ as before, but once this is done, we mark all bricks that are aligned under this transformation and remove them from their respective assemblies.  We then repeat this process with the remaining bricks and count how many rigid alignments must be computed until either the scene is empty, or the remaining bricks cannot be aligned because their shapes or colors do not match.  We then add an additional edit penalty of 1, representing a single edit to remove the brick, for each brick in $\hat{a}$ left at the end of this process, and a penalty of 2, representing an edit to add the brick to the assembly and an edit to move it into place, for each brick in $a^*$ left at the end of the sequence.

\textbf{$\mathbf{F1_e}$ score:} An added bonus of the \textbf{AED} metric is that it can be used to compute a matching between each brick in the generated assembly $\hat{a}$ and the target assembly $a^*$.  This matching allows us to compute one final metric: an F1 score over edges ($F1_e$), or connections between two bricks.  We consider every pair of bricks that are connected to each other in the generated assembly $\hat{a}$ to be a true positive edge if both of those bricks have been matched to a brick in the target assembly $a^*$ and the matching bricks in the target assembly are also connected to each other.  Otherwise the connected pair is a false positive.  Any connected pair in the target assembly $a^*$ that is not matched in this way is a false negative.  Like $F1_b$ this metric can be considered necessary but insufficient, but again it is useful because it lets us characterize the errors made during the build process.  If the agent was not able to determine the correct spatial alignment of the bricks, but is able to connect the right bricks together, then it may tell us the agent is struggling with the precise placement necessary to align bricks correctly.  This is similar to a metric used in Visual Genome \cite{visualgenome}, but uses our iterative matching edit distance to compute assignment and has no action/attribute labels on individual edges.

\subsection{Dataset}
\label{subsec:dataset}
We provide two sources of scene files to train and evaluate agents on these tasks.  The first is a set of fan-made reproductions of official LEGO sets that have been uploaded to the Open Model Repository (OMR)~\cite{omr}, while the second is a set of randomly constructed models that we have generated with the \LTRON{} simulator.

The OMR contains \NMODELS{} files that are incredibly diverse, ranging in size from \MINBRICKS{} to \MAXBRICKS{} bricks.  The sets come from over fifty distinct product categories such as ``City," ``Castle," and ``Star Wars" that have been released over a span of several decades and use \NBRICKS{} distinct brick shapes.  These files have many properties in common with other naturally occurring data sources such as a long tail of increasingly rare bricks, and edge-cases that are difficult to model.  This is a blessing to researchers who are interested in building models that can handle complex data distributions, and a curse to those looking for quick progress.  In general these models are much larger than we are presently able to train on.  Both the mean and the median number of bricks in a scene is more than one hundred, while our experiments below show that current methods struggle with scenes containing only eight bricks.  In order to generate a large amount of training data with smaller scenes, we have sliced these models into compact connected components using the connection points to find groups of connected bricks.  In all cases we have used a master train/test split on the original files to inform the train/test  on all slices of those files.  Table~\ref{tab:dataset} shows the train test splits for these slices.  See the supplementary material for more details on the statistics, slicing procedure and cleaning process of this data.

In contrast to the OMR data above, our randomly generated models are constructed by iteratively selecting brick shapes and colors at random and attempting to connect them to other bricks using randomly selected compatible connection points.  This provides a much larger source of data that is in many ways easier to use for training, but unfortunately has many qualitative differences from the more natural OMR data.  For example OMR scenes with a similar number of bricks tend to be much more compact than our randomly generated files as a byproduct of the human designers' preferences for tightly fitting configurations.  Similarly, the OMR scenes exhibit more symmetry, and more high-level structure such as clearly identifiable walls and branching structures.  Despite these issues, this randomly constructed data is still very useful as a way to explore how the problem becomes easier as we reduce the number of brick shapes.

\begin{table}
\centering
\begin{tabularx}{0.75\columnwidth}{X|c|c|c}
 \hline
 Open Model Repository & Train Scenes & Test Scenes & Total Scenes \\
 \hline
 Original Scenes & 1360 & 367 & \NMODELS{} \\
 2 Brick Slices & 136072 & 2000 & 138072 \\
 4 Brick Slices & 61514 & 2000 & 63514 \\
 8 Brick Slices & 28094 & 2000 & 30094 \\
 \hline
 Random Construction & Train Scenes & Test Scenes & Total Scenes \\
 \hline
 2 Bricks & 50000 & 2000 & 52000 \\
 4 Bricks & 50000 & 2000 & 52000 \\
 8 Bricks & 50000 & 2000 & 52000 \\
 \hline
\end{tabularx}
\vspace{5pt}
\caption{Train/test split sizes for the Open Model Repository and our Randomly Generated Data.}
\label{tab:dataset}
\end{table}

\section{Methods}
\label{sec:method}
\subsection{Model}
Our StudNet models are based on the popular Transformer\cite{transformer} architecture.  In this model, the input images are first broken into $16\times16$ pixel tiles similar to the VIT architecture\cite{vit}.  The model then extracts features from each tile using a learned linear layer and two positional embeddings, one that encodes the tile's XY coordinates in the image and another that encodes the tile's frame id in the temporal sequence.  We unroll the XY coordinates of the image into a single one-dimensional coordinate space and concatenate the coordinates of the table image and the hand image so that a single index can be used to determine which image the tile belongs to and its 2D location.  These tile features are then fed into a transformer that uses GPT-style\cite{gpt} causal masking to prevent tokens that occur early in the sequence from paying attention to later tokens.

Transformer models notoriously require very large memory due to the $N^2$ attention mechanism that allows for long-range connectivity between tokens in the sequence.  In order to make this architecture tractable on the long sequences of tokens produced by \LTRON{}, we employ a simple but effective data compression technique: at each step we only include image tiles which have changed since the previous frame.  In the first frame, we also remove all tiles that contain only the solid background color.  Given that manipulating a single brick usually only changes a small portion of the image, this results in substantial savings.  In addition to the image tiles, we provide a token that specifies the current phase (\PHASEA{} or \PHASEB{}).

The \PROBLEM{} task requires an agent to take both discrete high-level actions as well as select low-level pixel locations to assemble and disassemble bricks.  We model this using five separate heads: a mode head that selects one of the primary action types (see Figure \ref{fig:action_space}) to take at each step, a shape selector head and color selector head that are used when picking up a new brick, and a table location and hand location heatmap that is used to select pixel locations for brick interaction.  The shape and color heads are linear layers that project from the transformer hidden dimension to the number of shapes and colors used in a particular experiment.
Unfortunately we cannot decode a dense heatmap for the pixel locations directly from the tokens coming out of the transformer encoder because our compression strategy throws many of these tokens away.  We experiment with two different decoder styles to address this issue.

The first, which we refer to as StudNet-A uses a separate transformer decoder layer.  This layer receives a dense positional encoding as the query tokens, and the output of the encoder as the key and value tokens resulting in a dense output.  Although some details differ, this is similar to the Perceiver IO\cite{perceiverIO} and MAE\cite{mae} models that do primary computation at a lower resolution and use cross-attention to expand to dense output when necessary.  We decode at $16\times16$ resolution and upsample to $64\times64$.

The second decoder, which we refer to as StudNet-B, feeds the input images through a small convolutional network to produce a $64\times64$ feature map for the table and a $24\times24$ feature map for the hand.  In our experiments we use the first layer of a Resnet-18\cite{resnet} for this.  Two additional heads, one for the table workspace and another for the hand workspace, compute a single feature from a per-frame readout token, and use dot-product attention with the convolutional feature map to produce a heatmap of click locations.

We compare these models against a convolution and LSTM baseline.  This model takes guidance from the ALFRED Dataset \cite{alfred} which similarly requires an agent to reason about high level actions as well as pixel-based selection.  In this network, the images from both the table and hand workspaces are fed through a Resnet-18 backbone~\cite{resnet}, and are then concatenated and passed to an LSTM.  The output of this LSTM is then decoded using five heads.  The first three produce the mode, shape and color actions.  The second and third heads tile the LSTM feature to match the shape of the table and hand resnet features, then upsample these with UNET-style~\cite{unet}/FPN~\cite{fpn}-style lateral connections from the image encoder to produce a dense feature that is used to select cursor locations.  In experiments we use two versions of this model, one trained from scratch, and another where the Resnet-18 backbone has been pretrained on a pixel-labeling task designed to densely predict brick shapes and colors.

\subsection{Training}
We train the models above using behavior cloning on offline sequences.  In order to generate these sequences, we have developed a visual planner that interfaces directly with \LTRON{}.  This planner uses hidden state information combined with rendered occlusion maps to reason about which bricks are currently visible in the scene and plan assembly and disassembly sequences accordingly.  While this information allows the planner to determine which bricks can be manipulated, it does not strictly guarantee that the visual information acquired during the planning process is enough to unambiguously resolve the full 3D structure of the scene, or correctly identify the shapes of every brick.  This is due to the fact that many brick shapes look identical to others when viewed from certain angles or under partial occlusion, and so it may be important to change the camera viewpoint or disassembly order to resolve these ambiguities.  Due to the large number of brick shapes, we have not attempted to exhaustively catalogue when and how these ambiguities arise for every combination of brick shapes.  Therefore the planner currently has no way of knowing when these conditions occur.

\begin{figure*}[t!]
  \centering
  \makebox[\columnwidth][c]{\includegraphics[width=1.0\textwidth]{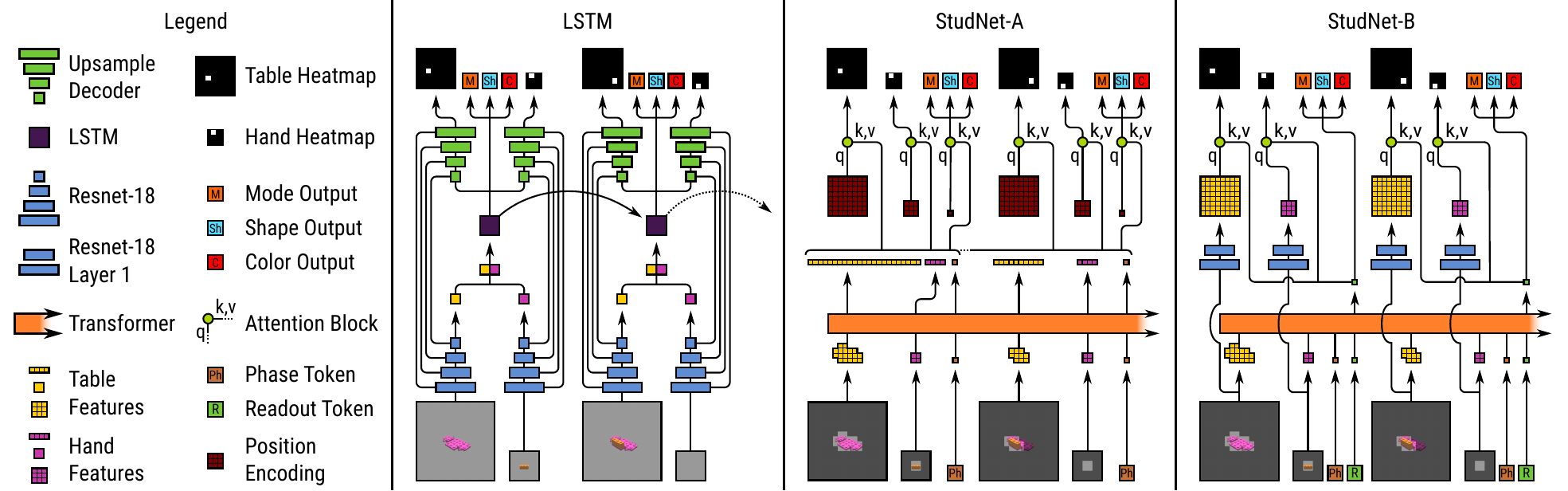}}
  \caption{Network architectures used in our experiments.}
  \label{fig:teaser}
  \vspace{-3mm}
\end{figure*}

\subsection{Limitations}
The visual planner can be quite slow and uses a two-stage process that requires reasoning over groups of individual actions.
Both of these issues make it difficult to use the planner as an expert for methods such as DAgger\cite{dagger} that require the expert to produce labels for sequences generated by the model. We therefore do not attempt to solve \PROBLEM{} using these approaches at the present time, and limit ourselves to methods that can train on a static dataset.  Building improved planners with the ability to quickly provide high-quality actions would be beneficial for this problem.

\section{Experiments}
\subsection{Break and Make}
\label{sec:experiments}
We evaluate the models above on the Random Construction and Sliced OMR datasets at three fixed scene sizes: two bricks, four bricks and eight bricks.  While these scenes are quite small compared to the complete models in the Open Model Repository, they often require dozens of interaction steps to complete and present a challenging problem.

On the random construction data with six brick types and six colors, all models make substantial progress on small scenes.  Table \ref{tab:random} shows the models' performance on each of these tasks under the four metrics described in Section~\ref{subsec:evaluation}.  Note that performance drops substantially as the scenes get larger.
\begin{table}
\centering
\begin{tabularx}{0.9\columnwidth}{Xcccc|cccc|cccc}
 \multicolumn{13}{c}{\textbf{Random Construction}} \\
  & \multicolumn{4}{c}{2 Bricks} & \multicolumn{4}{c}{4 Bricks} & \multicolumn{4}{c}{8 Bricks} \\
 \hline
 Metric & $F1_b$ & $F1_e$ & $F1_a$ & $AED$ & $F1_b$ & $F1_e$ & $F1_a$ & $AED$ & $F1_b$ & $F1_e$ & $F1_a$ & $AED$ \\
 \hline
 LSTM & 0.61 & 0.38 & 0.43 & 2.16 & 0.41 & 0.09 & 0.13 & 7.25 & 0.02 & 0.00 & 0.02 & 16.05 \\
 Pretr. LSTM & 0.70 & 0.51 & 0.45 & 1.89 & 0.25 & 0.01 & 0.08 & 8.46 & 0.03 & 0.00 & 0.02 & 16.09 \\
 StudNet-A & 0.90 & 0.86 & 0.58 & 1.11 & 0.56 & 0.29 & 0.24 & 5.80 & 0.02 & 0.01 & 0.01 & 15.87 \\
 StudNet-B & 0.87 & 0.77 & 0.57 & 1.30 & 0.64 & 0.34 & 0.25 & 5.48 & 0.38 & 0.14 & 0.12 & 13.90 \\
 \hline
\end{tabularx}
\vspace{5pt}
\caption{Test results of our four models on randomly constructed assemblies across three scene sizes.  See Section \ref{subsec:evaluation} for details on metrics.}
\label{tab:random}
\end{table}

The Sliced OMR dataset contains 1790 brick shapes and 98 colors making it structurally and visually significantly more challenging than the random construction dataset. Table \ref{tab:omr} illustrates that all of the models we tested score significantly lower on this dataset. In particular our StudNet-A transformer architecture fails to correctly learn to switch from disassembling to rebuilding the LEGO models and thus scores very poorly across all of our metrics. Our StudNet-B architecture shows the best overall performance, demonstrating that progress can be made even on the most challenging 8 brick dataset. This illustrates not only that  \PROBLEMLONG{} is a fundamentally hard problem, but also that its difficulty can be regulated by the dataset selection while maintaining the same action space and problem structure. This allows future work to make meaningful progress on simple datasets like Random Construction and then progress to ever more difficult datasets. 

\begin{table}
\centering
\begin{tabularx}{0.9\columnwidth}{Xcccc|cccc|cccc}
 \multicolumn{13}{c}{\textbf{Open Model Repository}} \\
  & \multicolumn{4}{c}{2 Bricks} & \multicolumn{4}{c}{4 Bricks} & \multicolumn{4}{c}{8 Bricks} \\
 \hline
 Metric & $F1_b$ & $F1_e$ & $F1_a$ & $AED$ & $F1_b$ & $F1_e$ & $F1_a$ & $AED$ & $F1_b$ & $F1_e$ & $F1_a$ & $AED$ \\
 \hline
 LSTM & 0.43 & 0.33 & 0.31 & 2.76 & 0.10 & 0.03 & 0.07 & 7.67 & 0.01 & 0.00 & 0.01 & 16.01\\
 Pretr. LSTM & 0.45 & 0.34 & 0.33 & 2.86 & 0.04 & 0.01 & 0.03 & 8.16 & 0.00 & 0.00 & 0.00 & 15.97 \\
 StudNet-A & 0.00 & 0.00 & 0.00 & 3.99 & 0.00 & 0.00 & 0.00 & 8.08 & 0.00 & 0.00 & 0.00 & 16.01 \\
 StudNet-B & 0.36 & 0.18 & 0.29 & 3.74 & 0.14 & 0.02 & 0.12 & 8.30 & 0.05 & 0.00 & 0.04 & 16.05 \\
 \hline
\end{tabularx}
\vspace{5pt}
\caption{Test results of our four models on OMR assemblies across three scene sizes.  See Section \ref{subsec:evaluation} for details on metrics.}
\label{tab:omr}
\end{table}

\subsection{Ablations and Failure Analysis}
Given the relatively low performance of the models presented here on the break and make task, we also conducted several experiments designed to discover which part of this problem is most difficult for future research.
Appendix D.1 contains the details of these experiments.
We also attempt to pretrain a model on the randomly generated assemblies and fine-tune on the OMR data.  Appendix D.2 contains details.  Finally we also provide a human baseline to verify the tractability of this problem in Appendix D.3.

\section{Conclusion}
\LTRON{} and the \PROBLEMLONG{} challenge offer an ideal environment to study a number  of important technical problems in Machine Learning and Artificial Intelligence.  First, the \LTRON{} simulator offers an environment to explore interactive building and construction problems at a level of detail and granularity that has not previously been possible.
Second, while we have only been able to make progress on very small LEGO models in this paper, \LTRON{} has the ability to represent very large assemblies with hundreds and even thousands of bricks.  Our hope is that the existence of these very difficult large-scale tasks that are currently beyond the scope of modern temporal-spatial visual modelling techniques will inspire researchers to explore new ways to scale algorithms and hardware to accomplish the goals.
Finally \PROBLEMLONG{} provides an ideal setting to explore interactive learning algorithms designed for long-term credit assignment, as agents must connect low-level actions taken during disassembly and inspection with reward signals collected in the distant future during reassembly.

\newpage



%
%
\bibliographystyle{splncs04}
\bibliography{refs}

\newpage

\section*{Appendix}

\appendix

\section{Qualatative Examples}

\begin{figure*}[b!]
  \centering
  \makebox[\columnwidth][c]{\includegraphics[width=0.82\textwidth]{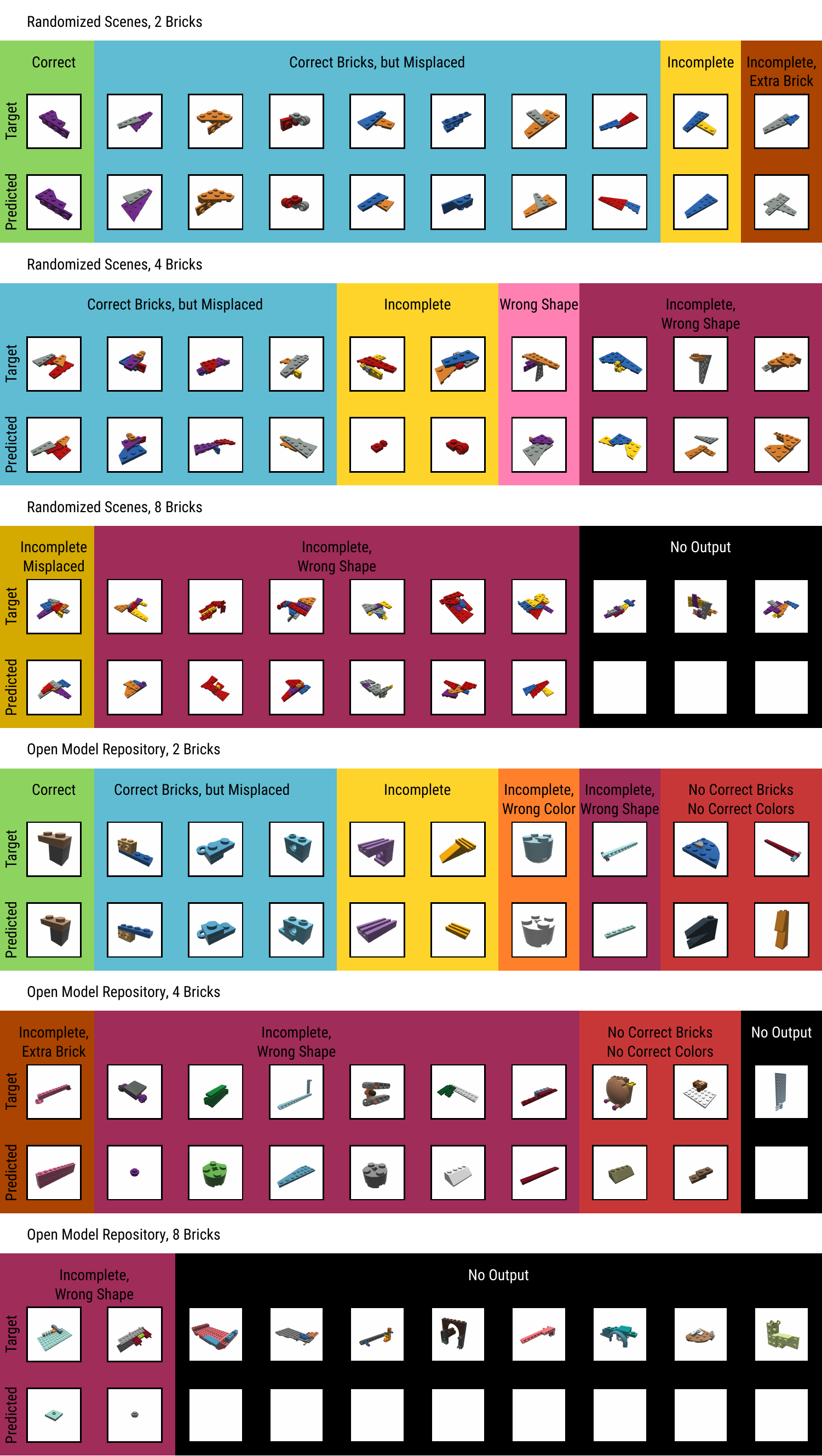}}
  \caption{Qualatative examples.}
  \label{fig:qualatative_examples}
  \vspace{-3mm}
\end{figure*}

Figure \ref{fig:qualatative_examples} shows ten randomly chosen target assemblies and assemblies predicted by the Stubnet-B model on each dataset.  Very few examples are completed precisely correct.  In the random 2-brick, 4-brick and the OMR 2-brick examples, the agent is able to build an assembly using the correct bricks but fails to connect them correctly.  The model largely fails to build anything resembling the target for the random 8-brick and OMR 4-brick and 8-brick scenes.  This clearly demonstrates the difficulty of this problem even in simple settings, and shows how the problem becomes more difficult as the number of bricks per scene increases, and the total number of bricks used in the dataset increases.  See the supplemental video for an example of a successful episode on a 2 brick model generated with the Stubnet-B model.

\label{subsec:dataset}

\section{Dataset}
\label{app:dataset}

\subsection{Simulator Details and Performance}
The \LTRON{} simulator uses native Python along with the Splendor-Render rendering package that provides a python interface for OpenGL rendering.  The speed of the simulator is dependent on the size and complexity of the scene, but for the scenes used in the experiments here, the simulator runs at over 100fps on an Nvidia 2080ti graphics card, and can be parallelized.  The simulator also supports headless EGL rendering for use on clusters that do not have graphical sessions.  \LTRON{} is not as heavily optimized as some other 3D environments such as Habitat 2.0 \cite{szot2021habitat}, but the simulator speed was fast enough not to be a bottleneck during experiments.

\subsection{Statistics}
We source brick shapes from the LDRAW\cite{ldraw} database, which contains over ten thousand official bricks.  However, many bricks in the official parts list are not used in any of the models in the OMR dataset.  Thus while \LTRON{} supports over ten thousand individual bricks, only around four thousand are used in the files that we slice to produce our training data.

\begin{figure*}[ht!]
  \centering
  \makebox[\columnwidth][c]{\includegraphics[width=1.0\textwidth]{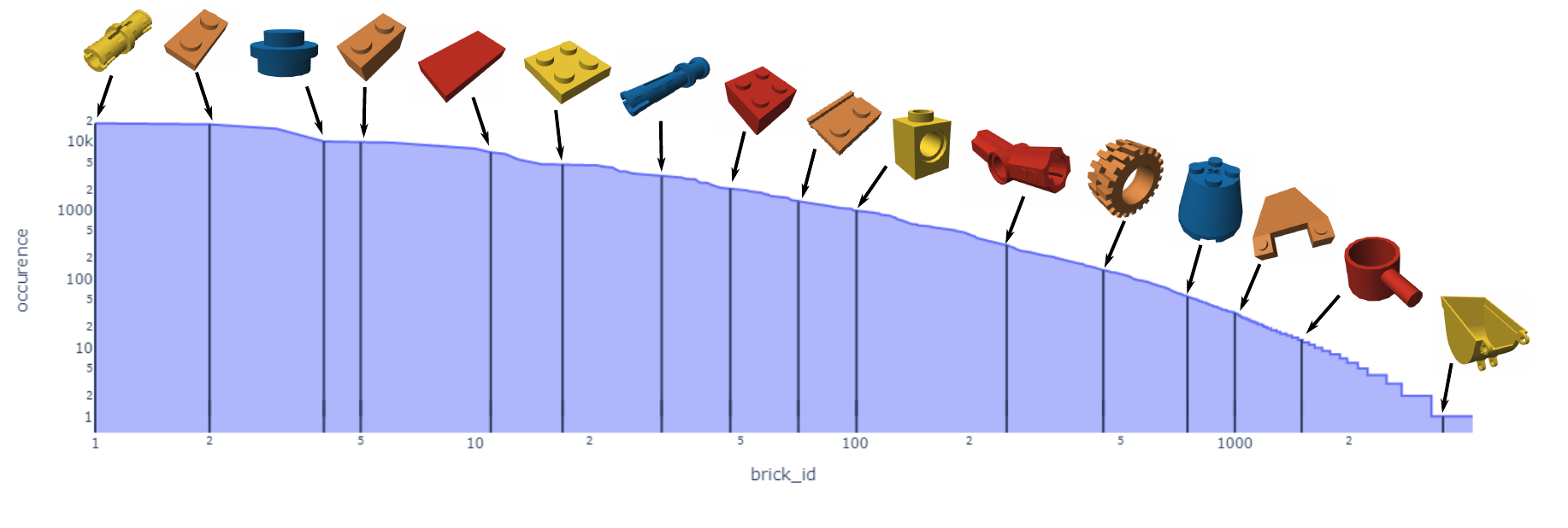}}
  \caption{Distribution of brick frequency in the OMR data with examples of various common and rare bricks.  The x-axis is the log-rank of each brick shape sorted by frequency with the most common brick on the left and the least common on the right.  The y-axis is the log-frequency of each class.  The most common brick is a simple connector pin which is used over 18,000 times throughout the dataset.  There are around one hundred classes with one thousand or more examples, and over five hundred classes with one hundred or more examples.  There is also a long tail of rare bricks, with more than half of all classes occurring less than ten times.}
  \label{fig:unigrams}
  \vspace{-3mm}
\end{figure*}

As one might expect, the distribution of part usage has a long tail of increasingly rare brick shapes.  To illustrate this, we plot the log of the individual brick usage counts against the log of the rank of these counts.  These diagrams are frequently used in language applications to demonstrate long-tail distributional statistics.  Figure \ref{fig:unigrams} illustrates the usage distribution.

\begin{figure}[h!]
  \centering
  \makebox[\columnwidth][c]{\includegraphics[width=0.7\columnwidth]{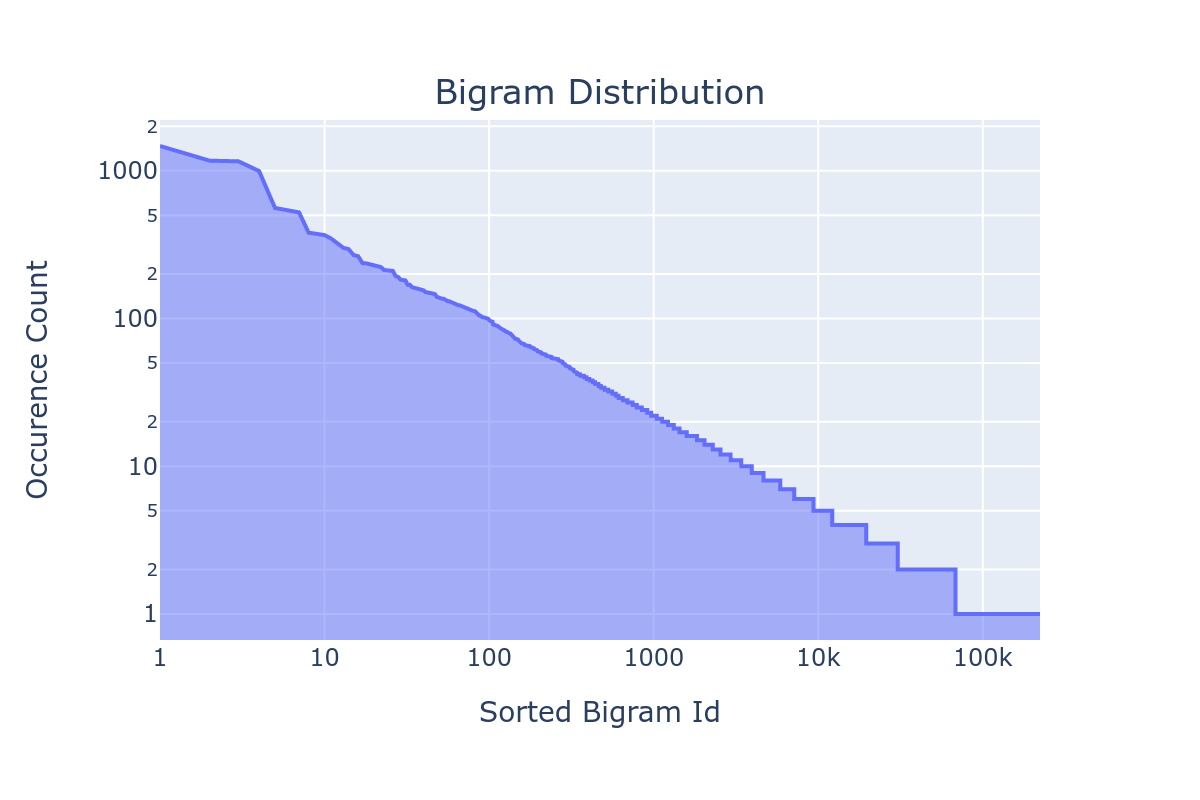}}
  \caption{The distribution of brick bigrams--two brick types and an associated transform--in the data included in \LTRON{}.}
  \label{fig:bigrams}
  \vspace{-3mm}
\end{figure}

In order to measure how common various combinations are we also compute statistics of brick \textit{bigrams} which consist of two bricks that are attached to each other with the same relative transform between them.  In total we have 222,679 bigrams where the most common one occurs 1,471 times in the data.  The bigrams also have a long tail of increasingly rare combinations as seen in Figure \ref{fig:bigrams}.

\subsection{Data Preprocessing}

When working with data from the Open Model Repository, we have taken several steps to make the data more manageable for the learning agent.

First we have blacklisted several bricks that are very large that would not fit into the default screen viewport.

\begin{figure}[h!]
  \centering
  \makebox[\columnwidth][c]{\includegraphics[width=0.5\columnwidth]{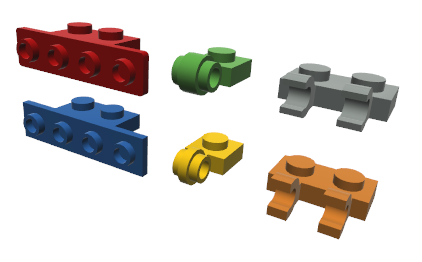}}
  \caption{Examples of brick variants in the LDRAW repository.  Note the red brick has slightly rounded corner compared to the blue brick.  The thick section between the two connection points is slightly thicker in the green brick than the yellow.  The shape of the clips is slightly thinner in the grey brick than the orange.  As a preprocessing step we replaced each of these categories with a single canonical version when using these bricks for \PROBLEM{}.}
  \label{fig:variants}
  \vspace{-3mm}
\end{figure}

Second, many of the brick shapes in the LDRAW repository represent small changes and variations that have been made to different bricks over the years.  Where possible, we have collapsed multiple brick shapes that represent only minor variations into a single class by replacing variations with a single canonical version of a part.  This further reduced the number of shapes in our dataset from over four thousand to \NBRICKS{}.  Figure \ref{fig:variants} shows examples of bricks bricks that have been collapsed into a single category.

Third, for each model in the Open Model Repository, we have computed all connected components in the model and split out each as a separate file.  These connected components represent groups of bricks that are all attached to each other using the connection points that \LTRON{} currently supports.  This is important as many official LEGO sets contain multiple detached components, for example a single racing set may contain two separate cars.

Finally, once we have these connected components, we slice them into smaller components in order to generate many examples of small scenes for training.  When slicing we aim to keep the models as small and compact as possible, so we use a simple greedy algorithm that selects the first brick in the scene, then looks at other bricks it is connected to and selects the one that minimizes the largest axis of the bounding box of the new model, and repeats this process until the desired number of bricks has been selected.  These bricks are then broken out as their own new model file and removed from the scene.  Then this process is repeated until nothing remains of the original model.  This process ensures that the resulting sliced models have similar distributional statistics as the original data, not only in terms of individual brick usage, but also local neighborhood structure.

\begin{figure*}[b!]
  \centering
  \makebox[\columnwidth][c]{\includegraphics[width=1.0\textwidth]{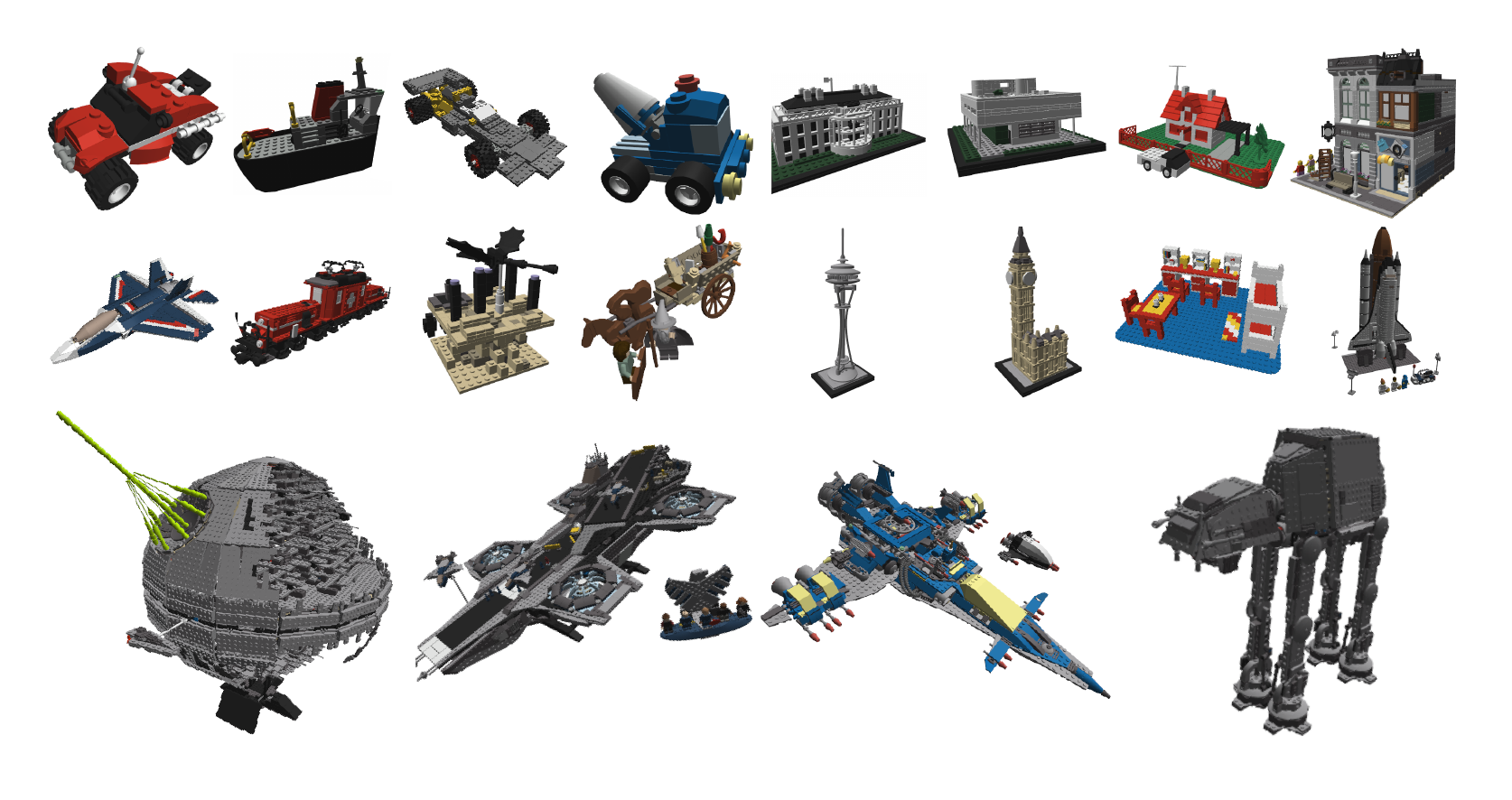}}
  \caption{Several examples of models from the Open Model Repository.  Each model in the bottom row has over eight hundred bricks, each of which can be individually manipulated within LTRON.}
  \label{fig:teaser}
  \vspace{-3mm}
\end{figure*}
\label{subsec:dataset}

\subsection{Symmetry}
Many LEGO bricks exhibit rotational symmetry about a one of the three primary X, Y, Z axes.  We have generated a table describing these symmetries by automatically analyzing the shape of each brick in order to take symmetry into account when scoring the final models.  In order to compute this table, we render a depth map of each brick from six canonical directions (+/- X,Y,Z).  We then rotate each brick by 90, 180 and 270 degrees about each of the X, Y and Z axes and re-render these depth maps.  We mark any transformation that produces approximately identical depth maps as a symmetry.

\section{Ablations and Analysis}
Given the relatively low performance of the models presented here on the break and make task, we also conducted several experiments designed to discover which part of this problem is most difficult.

\subsection{Ablations}
We first modified the environment to provide the correct pick actions (selection of brick shape and color) automatically when necessary.  If numbers on this experiment improved dramatically, this would indicate that the models were having difficulty remembering the brick shapes and colors that were observed during the \PHASEA{} phase.  For the sake of space, we ran this experiment on the randomly constructed 2-brick assemblies using the StudNet-B model.  As shown in table \ref{tab:gt_insert} this yields a very small performance gain. This result, taken together with the fact that the $F1_b$ score strongly outperforms the $F1_a$ score in almost all experiments indicates that remembering the brick shape and color, and learning when to insert them is not a primary source of error.

We also extracted frames from the random 2-brick assemblies and trained a single-frame FCOS\cite{fcos} detection model with additional heads to predict 3D position and orientation.  This is designed to determine if estimating the 3D pose of the bricks in a single frame is a possible point of failure.  To evaluate this model, we use an AP score where we consider an estimated brick to be a true-positive match with a ground truth brick if their shapes and colors match, and their poses are within 30 degrees and 8mm of each other.  In this setting a ResNet50 backbone scores 0.97 AP, a ResNet18 backbone scores 0.96 AP and a transformer backbone scores 0.81 AP.  These results indicate that estimating the identities and poses of the bricks in a single frame is also not a major challenge.

Taken together, these two results provide indirect evidence for the hypothesis that the most challenging part of this problem is the need for spatially-precise long term memory, and a large interactive action space.

\begin{table}
\centering
\begin{tabularx}{0.9\columnwidth}{Xcccc}
 \multicolumn{5}{c}{\textbf{Ground Truth Insertion}} \\
 \hline
 Random 2 Brick & $F1_b$ & $F1_e$ & $F1_a$ & $AED$\\
 \hline
 StudNet-B Original & 0.87 & 0.77 & 0.57 & 1.30\\
 StudNet-B GT-Insert & 0.88 & 0.84 & 0.57 & 1.12\\
 \hline
\end{tabularx}
\vspace{5pt}
\caption{Test results when providing ground truth brick insertion operations (bottom) compared to original performance (top).}
\label{tab:gt_insert}
\vspace{-15pt}
\end{table}

\subsection{Fine Tuning}
We also conducted an experiment to test whether a network trained on random construction assemblies could be fine-tuned to the OMR assemblies.  Due to the mismatched number of brick shapes and colors, we trained new color and shape heads for the OMR data.  As shown in table \ref{tab:finetune} performance actually gets slightly worse, except for edit distance, which improves slightly.

\begin{table}
\centering
\begin{tabularx}{0.9\columnwidth}{Xcccc}
 \multicolumn{5}{c}{\textbf{Pretrain/Fine Tune}} \\
 \hline
 OMR 2 Brick & $F1_b$ & $F1_e$ & $F1_a$ & $AED$\\
 \hline
 StudNet-B Original & 0.36 & 0.18 & 0.29 & 3.74\\
 StudNet-B Finetune & 0.29 & 0.11 & 0.25 & 3.32\\
 \hline
\end{tabularx}
\vspace{5pt}
\caption{Test results when using a model pretrained on the Random-2 dataset and fine-tuning on OMR-2 (bottom) compared to training from scratch on OMR (top).}
\label{tab:finetune}
\vspace{-15pt}
\end{table}

\subsection{Human Baseline}
In order to make sure the \PROBLEM{} task is possible using the interface provided, we also conducted a small user-experiment.  Using a rudimentary interface, our first author was able to perfectly reconstruct 8 out 10 randomly sampled scenes from the Random Construction 2 dataset.  The two failures contained a single placement mistake each which could not be fixed within the maximum episode length. This would yield F1b: 1.0, F1e: 1.0, F1a: 0.90 and an AED: 0.2, which is far better than any of the baseline models.  This shows that the task is feasible, and that there is substantial room for improvement in the approaches here.

\section{Ethical Research and Intellectual Property}
\label{app:ethics}
The \LTRON{} environment makes substantial use of data collected from the internet.  Whenever working with data that comes from external sources, it is important to protect individuals' privacy and respect the intellectual property rights of the original authors.

To our knowledge, \LTRON{} does not pose a substantial privacy risk to individuals.  The only personally identifiable information contained in \LTRON{} is a name or pseudonym of the individual author of a particular file.  These authors have chosen to make their names public, and have requested attribution through Creative Commons licensing when using their work.

Intellectual Property is an important consideration for this project as each of the files included in \LTRON{} represent a substantial investment of time and effort by the original authors.  We have therefore only included files that have clear and unambiguous open licensing terms.  Fortunately the LDraw Open Model Repository \cite{ldraw} contains over a thousand high-quality files with generous Creative Commons licensing.  See Appendix \ref{app:acknowledgements} for authorship details.  \LTRON{} is inter operable with other user-generated content that can be found scattered throughout various community forums.  In most cases these do not contain explicit licensing terms, so we have not included this data for distribution and have not used these files in our experiments.  In the future, we may augment \LTRON{} with more data when and if we are able to secure open licensing agreements with individual authors.

LEGO is an official trademark of the LEGO Group which is not affiliated with this paper or the authors, and has not endorsed or sponsored this paper.  To our knowledge, all material in \LTRON{} has been generated either by the fan community or this paper's authors, and are therefore not under copyright or other intellectual property protection by the LEGO Group.

\section{Acknowledgements}
\label{app:acknowledgements}
The following is a list of contributors to the LDraw project \cite{ldraw}, the Open Model Repository and the LDCad metadata \cite{ldcad}.  These online resources have been created and are maintained by a large group of volunteers.  These contributors are not affiliated with, and have not directly contributed to the authorship of this paper, however they have provided open resources which have been of enormous value to the \LTRON{} environment.  We include their names here in order to acknowledge their efforts and contribution.

\textbf{LDraw Part Authors:}
These authors have contributed parts or parts of parts to the primary LDraw database.  This database contains the geometric shape of over 13,000 individual brick types, and is used in \LTRON{} as the primary source of renderable geometry.  The authors are listed here in the order of the number of authored files:

\textit{
Philippe Hurbain [Philo],
Magnus Forsberg [MagFors],
James Jessiman,
Steffen [Steffen],
Chris Dee [cwdee],
Michael Heidemann [mikeheide],
J.C. Tchang [tchang],
Gerald Lasser [GeraldLasser],
Alex Taylor [anathema],
Guy Vivan [guyvivan],
Tore Eriksson [Tore\_Eriksson],
Willy Tschager [Holly-Wood],
Max Martin Richter [MMR1988],
Damien Roux [Darats],
Andy Westrate [westrate],
Christian Neumann [Wesley],
Ulrich R der [UR],
Steve Bliss [sbliss],
Franklin W. Cain [fwcain],
Rolf Osterthun [Rolf],
Massimo Maso [Sirio],
Santeri Piippo [arezey],
Marc Klein [marckl],
Niels Karsdorp [nielsk],
Evert-Jan Boer [ejboer],
Johann Eisner [technicbasics],
Donald Sutter [technog],
Vincent Messenet [Cheenzo],
Joerg Sommerer [Brickaneer],
Paul Easter [pneaster],
Daniel Goerner [TK-949],
Bernd Broich [bbroich],
Nils Schmidt [BlackBrick89],
Stan Isachenko [angmarec],
William Howard [WilliamH],
John Van Zwieten [jvan],
Howard Lande [HowardLande],
Thomas Burger [grapeape],
Owen Burgoyne [C3POwen],
Orion Pobursky [OrionP],
Takeshi Takahashi [RainbowDolphin],
Mikkel Bech Jensen [gaia],
Arne Hackstein,
John Riley [jriley],
Tim Gould [timgould],
Mark Kennedy [mkennedy],
Tony Hafner [hafhead],
Kevin Roach [KROACH],
Tim Lampmann [L4mpi],
Greg Teft [gregteft],
Christophe Mitillo [Christophe\_Mitillo],
Leonardo Zide,
Sylvain Sauvage [SLS],
Merlijn Wissink [legolijntje],
Mark Chittenden [mdublade],
Carsten Schmitz [Deckard],
Jaco van der Molen [Jaco],
Luis E. Fernandez [lfernand],
Miklos Hosszu [hmick],
Sascha Broich,
Lutz Uhlmann [El-Lutzo],
Ronald Vallenduuk [Duq],
Matt Schild [mschild],
Thomas Woelk [t.woelk],
Manfred Moolhuysen,
Ross Crawford [rosco],
Bertrand Lequy [Berth],
Imre Papp [ampi],
Remco Braak [remco1974],
George Barnes [glbarnes],
Jude Parrill [theJudeAbides],
Marc Giraudet [Mad\_Marc],
Ludo Soete [ludo],
El'dar Ismagilov [Eldar],
Bjoern Sigve Storesund [Storesund],
Tomas Kralicek [RabbiT\_CZ],
John Troxler [Gargan],
Jonathan Wilson [jonwil],
Jeff Boen [onyx],
Lutz Uhlmann,
Peter Watts [FrozenPea],
Niels Bugge [SirBugge],
Marc Schickele [samrotule],
Alexandre Bourdais [x-or],
Jeff Boen,
Ingolf Weisheit [stahlwollschaf],
Andrew Ananjev [woozle],
Damien Guichard [BrickCaster],
Heiko Jelnikar [KlotzKiste],
Yann Bouzon [Zaghor],
Stephan Meisinger [smr],
Lee Gaiteri [LummoxJR],
Marek Idec [Maras],
Lars C. Hassing [larschassing],
Nathan Wright,
Remco Braak,
Kevin Clague [kclague],
Larry Pieniazek [lar],
Matthew J. Chiles [mchiles],
Guus-Jan Wijnhoven [guus],
Victor Di Rienzo [tatubias],
Shimpei Ohsumi [Shimpei-Ohsumi],
Christian M. Angele [cma\_1971],
Sven Moritz Hein [smhltec],
Ildefonso Zanette [izanette],
Ronald Scott Moody [rmoody],
Paul Izquierdo Rojas [pir],
Joseph H. Cardana,
Taylor Bangs [DoomTay],
Peter Lind [peterlinddk],
Adriano Aicardi,
Jonathan P. Brown,
Karim Nassar,
Dee Earley [DeannaEarley],
Brent Jackson [bjackson],
Lance Hopenwasser [cavehop],
Bram Lambrecht,
Ishino Keiichiro,
Joachim Probst,
Yu Zhang [ishkafel],
David Manley [djm],
Joshua Delahunty [dulcaoin],
Ignacio Fernandez Galvan [Jellby],
Heather Patey,
Adam Howard [Whist],
N. W. Perry [Plastikean],
Matthew Morrison [cuddlyogre],
Dennis Osborn,
Robert Sexton [rhsexton],
Sybrand Bonsma [Sybrand],
Remco Canten [rempie],
R. M. Rodinsky [dublar],
Paul Schelleman [schellie],
Bob LeVan [kyphurious],
Dave Schuler,
Jens Bauer [rockford],
Jan Folkersma [Stinky],
Graham Wilkes [remorse],
Damien Duquennoy,
Svend Eisenhardt [eisenhardt],
Enzo Silvestri [ienzisolves],
Bert Van Raemdonck [BEAVeR],
Andreas Weissenburg [grubaluk],
Amnon Silverstein [Amnon],
Frits Blankenzee,
Zoltan Keri [kzoltan82],
Alex Forencich [aforencich],
Daniele Benedettelli [benedettelli],
Rafael Skibicki [Rola],
Carlos Arbesu [NXTbesu],
Philip Peickert [mr51flip],
Sam Roberts [sroberts],
Gene Welborn [dtaax],
Allister Mclaren [amclaren],
Joao Almeida [TullariS],
Derrick Chiu [LordAdmiral],
Niklas Buchmann [NiklasB],
Alex Seeley [alex],
Paolo Campagnaro [pcampagn],
Ryan Dennett,
Maciej Kowalik [Madmaks],
Ian Reid [Ian\_Reid],
Jeff Stembel,
C.L.Rasmussen [johnny-thunder],
Don Heyse,
Chris Moseley,
Steve Chisnall [StevieC],
Edwin Pilobello [gypsy\_fly],
John Boozer [jediknight219],
Ben Lyttle [legotrek],
Reinhard "Ben" Beneke [Ben\_aus\_BS],
Jim DeVona [anoved],
Jason Mantor [Xanthra47],
James Mastros [theorbtwo],
Jeffery MacEachern [legonerd],
Axel Poque,
John Jensen,
Douglas Taylor, Jr. [djcool905],
Stig-Erik Blomqvist [stigge72],
Jeroen Ottens,
Bernd Munding,
Steve Demlow [demlow],
Bert J. Giesen,
Reuben Pearse [ReubenPearse],
Robert Paciorek [bercik],
Richard Baxter [rbaxter],
Dan Boger [dan],
Duane Hess,
Earnest J. Banbury [Banbury],
Martin G Cormier,
Jack Hawk [jhawk],
Arthur Sigg [Pendulum],
James Shields [lostcarpark],
Richard Finegold,
Martyn Boogaarts,
Antony Caparica [antonyc],
Christopher Bulliner [CMB27],
Christopher Pedersen [pedersen],
Troy [peloquin],
Travis Cobbs [tcobbs].
}

\textbf{LDRAW Open Model Repository Authors:}
These authors have contributed full models to the LDraw Open Model Repository.  This repository contains 1,727 high quality models assembled from the bricks in the LDraw parts library.  Again the authors are listed in the order of the number of files contributed:

\textit{
Robert Paciorek [bercik],
Philippe Hurbain [Philo],
Marc Giraudet [Mad\_Marc],
Massimo Maso [Sirio],
Damien Roux [Darats],
Merlijn Wissink [Legolijntje],
Willy Tschager [Holly-Wood],
Orion Pobursky [OrionP],
Max Martin Richter [MMR1998],
Stefan Frenz [smf],
Tomas Kralicek [RabbiT\_CZ],
Victor Di Rienzo [tatubias],
Ken Drew [Ken],
Johann Eisner [technicbasics],
Bert Van Raemdonck [BEAVeR],
Evert-Jan Boer [ejboer],
Charles Farmer [farmer],
Marc Belanger [MonsieurPoulet],
Roland Dahl [RolandD],
Oh-Seong KWON,
Jude Parill [theJudeAbides],
Ignacio Fernandez Galvan [jellby],
Faramond Florent [Makou],
Daniel Goerner [TK-949],
Zoltán Kéri [Zoltank82],
Takeshi Takahashi [RainbowDolphin],
Stan Isachenko [angmarec],
Jaco van der Molen [Jaco],
Rijk van Voorst [Rijkjavik],
Christian Maglekær,
Christian Neumann [Wesley],
Steffen [Steffen],
N. W. Perry [Plastikean],
Michael Heidemann [mikeheide],
Michal Oravec [Bloodybeast],
Magnus Forsberg [MagFors],
Lasse Deleuran [Lasse Deleuran],
[juraj3579],
Allard van Efferen [aefferen],
Rafael Skibicki [Rola],
Adrien Pennamen [AdrienPennamen],
Greg Teft [gregteft],
Christophe Mitillo [Christophe\_Mitillo],
René Frijhoff,
Sean Burke [Leftmost],
Antony Lodge,
Mirjan Lipovcan [LegoZG],
Florian Schüller [schuellerf],
Tim Singer [tsinger],
[EdmanZA],
Guido Mauro,
Casey Puyleart,
Peter Bartfai,
Oliver Damman [Nautilus],
Steffen Altenburg [SteffenA],
Ulrich Röder [UR],
Alexey [folkoluck],
Kevin Hendirckx [Gebruiker].}

\textbf{LDCAD Metadata Authors:}
\LTRON{} uses LDraw metadata bundled with the free LDCAD software to find connection points between bricks.  \textit{Roland Melkert} is the primary author of LDCAD, but others have contributed to this metadata.  They are listed here, again in order of the number of files contributed:

\textit{
Roland Melkert [roland],
Philippe Hurbain [Philo],
Milan Vancura [MilanV],
Alex Taylor [anathema],
Jason McReynolds [Jason McReynolds].
}

\textbf{Additional Acknowledgements:}
We also wish to acknowledge \textit{Toby Nelson}, the author of the open-source \textit{ImportLDraw} plugin for Blender.  This was used to convert the LDraw parts library to wavefront OBJ files for use in our rendering system.

\end{document}